\documentclass[sigconf]{acmart}

\AtBeginDocument{%
  \providecommand\BibTeX{{%
    \normalfont B\kern-0.5em{\scshape i\kern-0.25em b}\kern-0.8em\TeX}}}

\usepackage{graphicx}
\usepackage{amsmath}
\usepackage{booktabs}
\usepackage{rotating}
\usepackage{xfrac}

\usepackage{hyperref}

\usepackage[acronym]{glossaries}
\makeglossaries
\newacronym{bnns}{BNN's}{Binary Neural Networks}
\newacronym{qnns}{BNN's}{Quantized Neural Networks}
\newacronym{nlp}{NLP}{natural language processing}

\newacronym{sbt}{SBT}{Sparse Binary Transformer}
\newacronym{pot}{POT}{Peak Over Threshold}

\newacronym{flops}{FLOPs}{floating point operations}

\newacronym{lstf}{LSTF}{long-term time series forecasting}

\newacronym{fp32}{FP32}{floating point 32}

\usepackage{epsfig}
\usepackage{comment}
\usepackage{graphicx}
\usepackage{amsmath}
\usepackage{booktabs}
\usepackage{makecell}
\usepackage{multirow}
\usepackage{bm}
\usepackage{arydshln}

\AtBeginDocument{%
 \providecommand\BibTeX{{%
  Bib\TeX}}}

\copyrightyear{2023}
\acmYear{2023}
\setcopyright{rightsretained}
\acmConference[KDD '23]{Proceedings of the 29th ACM SIGKDD Conference on Knowledge Discovery and Data Mining}{August 6--10, 2023}{Long Beach, CA, USA}
\acmBooktitle{Proceedings of the 29th ACM SIGKDD Conference on Knowledge Discovery and Data Mining (KDD '23), August 6--10, 2023, Long Beach, CA, USA}
\acmDOI{10.1145/3580305.3599508}
\acmISBN{979-8-4007-0103-0/23/08}

\begin{document}

\title{Sparse Binary Transformers for Multivariate Time Series Modeling}

\author{Matt Gorbett}
\email{matt.gorbett@colostate.edu}
\affiliation{%
  \institution{Colorado State University}
  \city{Fort Collins, CO}
  \country{USA}
}

\author{Hossein Shirazi}
\email{hshirazi@sdsu.edu}
\affiliation{%
  \institution{San Diego State University}
  \city{San Diego, CA}
  \country{USA}
}

\author{Indrakshi Ray}
\email{indrakshi.ray@colostate.edu}
\affiliation{%
  \institution{Colorado State University}
  \city{Fort Collins, CO}
  \country{USA}
}


\begin{abstract}
Compressed Neural Networks have the potential to enable deep learning across new applications and smaller computational environments.  However, understanding the range of learning tasks in which such models can succeed is not well studied.  
In this work, we apply \textit{sparse} and \textit{binary-weighted} Transformers to multivariate time series problems, showing that the lightweight models achieve accuracy comparable to that of dense floating-point Transformers of the same structure.
Our model achieves favorable results across three time series learning tasks: \textit{classification}, \textit{anomaly detection}, and \textit{single-step forecasting}.  
Additionally, to reduce the computational complexity of the attention mechanism, we apply two modifications, which show little to no decline in model performance: 1) in the classification task, we apply a fixed mask to the query, key, and value activations, and 2) for forecasting and anomaly detection, which rely on predicting outputs at a single point in time, we propose an attention mask to allow computation only at the current time step.
Together, each compression technique and attention modification substantially reduces the number of non-zero operations necessary in the Transformer. 
We measure the computational savings of our approach over a range of metrics including parameter count, bit size, and floating point operation (FLOPs) count, showing up to a $53\times$ reduction in storage size and up to $10.5\times$ reduction in FLOPs.

\end{abstract}





\maketitle

\section{Introduction}
The success of deep learning can largely be attributed to the availability of massive computational resources \cite{krizhevsky_imagenet_2012, simonyan_very_2014,he_deep_2016}.
Models such as the Transformer \cite{vaswani_attention_2017} have changed machine learning in fundamental ways, producing state-of-the-art results across fields such as \gls{nlp}, computer vision \cite{chen_pre-trained_2021,touvron_training_2021}, and time series learning \cite{zerveas_transformer}.
Much effort has been aimed at scaling these models towards \gls{nlp} efforts on large datasets \cite{devlin_bert_2019,brown_language_2020}, however, such models cannot practically be deployed in resource-constrained machines due to their high memory requirements and power consumption. 



Parallel to the developments of the Transformer, the Lottery Ticket Hypothesis \cite{frankle_lottery_2019} demonstrated that neural networks contain sparse subnetworks that achieve comparable accuracy to that of dense models.
Pruned deep learning models can substantially decrease computational cost, and enable a lower carbon footprint and the democratization of AI.  
Subsequent work showed that we can find highly accurate subnetworks within randomly-initialized models without training them \cite{ramanujan_whats_2020}, including binary-weighted neural networks \cite{diffenderfer_multi-prize_2021}.
Such ``lottery-ticket'' style algorithms have mostly experimented with image classification using convolutional architectures, however, some work has shown success in pruning \gls{nlp} Transformer models such as BERT \cite{chen2020lottery,ganesh2021compressing,jiao-etal-2020-tinybert}. 

\begin{figure}
\includegraphics[width=8cm]{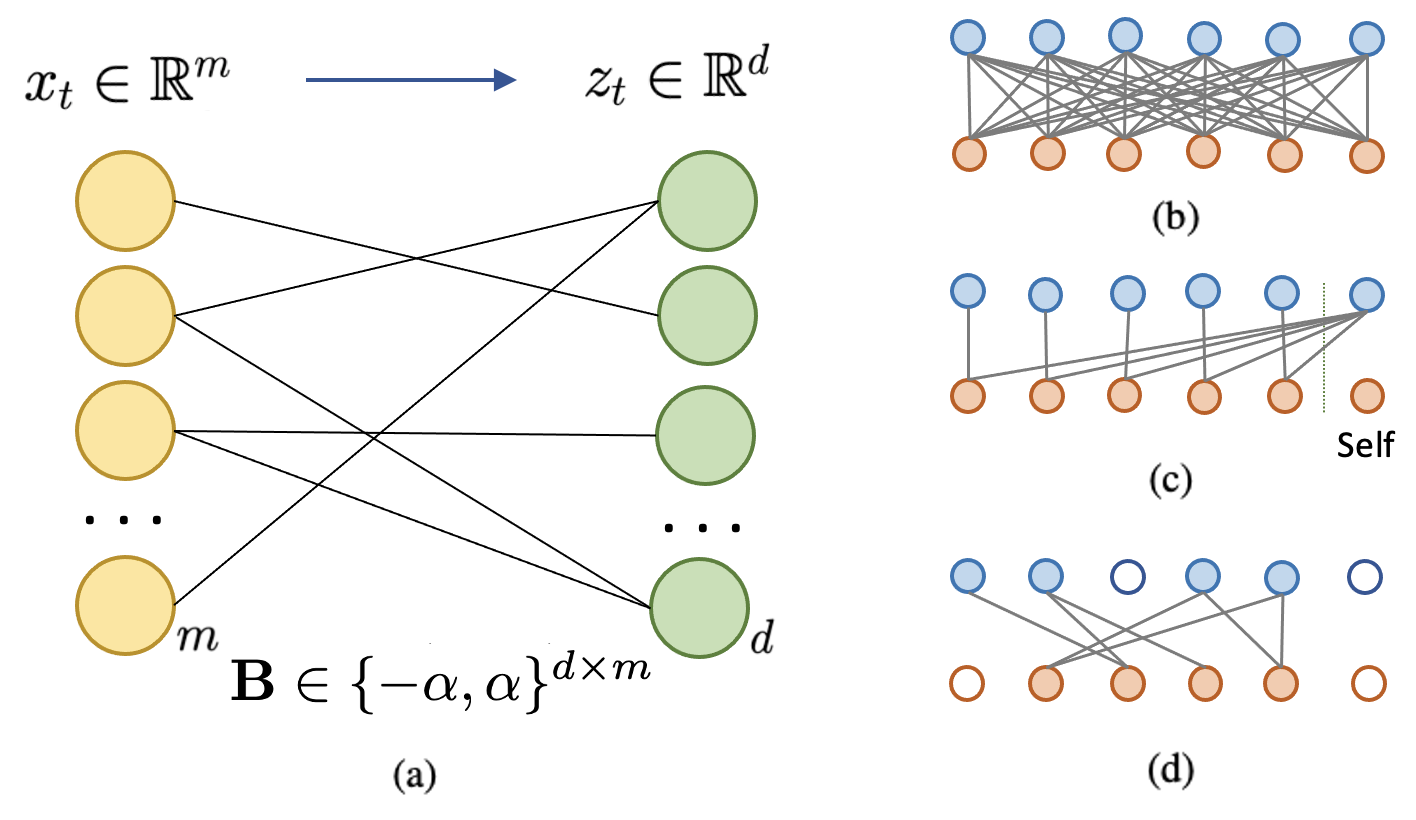}
\caption{A sparse binary linear layer (left) and various attention modules (right). a) An example of a sparse and binary linear module, with binary weights $\mathbf{B}$ scaled to $\{-\alpha, \alpha\}$.  b) A fully-connected attention module, where each point represents a time step ($w=6$). c) The Step-T attention module, where each past time point attends to itself and the latest time point $t$ attends to all past time points. d) An attention module with sparse Query (Q), Key (K), and Value (V) activations.  }
\label{teaser}
\end{figure}

In this work, we extend the Lottery Ticket Hypothesis to time series Transformers, showing that we can prune and binarize the weights of the model and still maintain an accuracy similar to that of a Dense Transformer of the same structure. 
To achieve this, we employ the Biprop algorithm \cite{diffenderfer_multi-prize_2021}, a state-of-the-art technique with proven success on complex datasets such as ImageNet \cite{5206848}.
The combination of weight binarization and pruning is unique from previous efforts in Transformer compression.
Moreover, each compression technique offers separate computational advantages: neural network pruning decreases the number of non-zero \gls{flops}, while binarization reduces the storage size of the model.
The Biprop algorithm's two compression methods rely on each other during the training process to identify a high-performing subnetwork within a randomly weighted neural network.
The combination of pruning and weight binarization is depicted in Figure \ref{teaser}a.

We apply our approach to multivariate time series modeling.
Research has shown that Transformers achieve strong results on time series tasks such as classification \cite{zerveas_transformer}, anomaly detection \cite{xu_anomaly_2022,tuli_tranad_2022}, and forecasting \cite{liu2021pyraformer, zhou2021informer}.
Time series data is evident in systems such as IoT devices \cite{cook2019anomaly}, engines \cite{malhotra2016lstm}, and spacecraft \cite{su_robust_2019,baireddy2021spacecraft}, where new insights can be gleaned from the large amounts of unmonitored information.  
Moreover, such systems often suffer from resource constraints, making regular deep learning models unrealistic -- for instance, in the Mars rover missions where battery-powered devices are searching for life \cite{bhardwaj2021semi}.
Other systems such as satellites contain thousands of telemetry channels that require granular monitoring.
Deploying large deep learning models in each channel can be extremely inefficient.
As a result, lightweight Transformer models have the potential to enhance a wide variety of applications.  

In addition to pruning and binarizing the Transformer architecture, we simplify the complexity of the attention mechanism by applying two modifications.
For anomaly detection and forecasting, which we model using overlapping sliding window inputs, we apply an attention mask to only consider attention at the current time step instead of considering attention for multiple previous time steps.
For classification tasks, we apply a static mask to the query, key, and value projections, showing that only a subset of activations is needed in the attention module to achieve the same accuracy as that obtained using all the activations.

Finally, we estimate the computational savings of the model in terms of parameters, storage cost, and non-zero \gls{flops}, showing that pruned and binarized models achieve comparable accuracy to dense models with substantially lower computational costs.  
\newline
Our contributions are as follows:
\begin{itemize}
  \item We show that sparse and binary-weighted Transformers achieve comparable accuracy to Dense Transformers on three time series learning tasks (classification, anomaly detection, forecasting).
  To the best of our knowledge, this is the first research examining the efficacy of compressed neural networks on time series related learning.  
  \item We examine pruning and binarization jointly in Transformer-based models, showing the benefits of each approach across multiple computational metrics.
  Weight binarization of Transformer based architectures has not been studied previously.   

\end{itemize}

These findings provide new potential applications for the Transformer architecture, such as in resource-constrained environments that can benefit from time series related intelligence.




\section{Related Work}\label{related}
In this section, we describe existing research related to Transformers in time series modeling, neural network pruning and compression, and finally efficient Transformer techniques. 

\subsection{Transformers in Time Series}
Various works have applied Transformers to time series learning tasks \cite{wen_transformers_2022}.
The main advantage of the Transformer architecture is the attention mechanism, which learns the pairwise similarity of input patterns.
Moreover, it can efficiently model long-range dependencies compared to other deep learning frameworks such as LSTM's \cite{liu2021pyraformer}.
Zerveas et al. \cite{zerveas_transformer} showed that we can use unsupervised pretrained Transformers for downstream time series learning tasks such as regression and classification.
Additional work in time series classification has proposed using a ``two tower" attention approach with channel-wise and time-step-wise attention \cite{liu2021gated}, while other work has highlighted the benefits of Transformers for satellite time series classification compared to both recurrent and convolutional neural networks  \cite{russwurm2020self}.  

For anomaly detection tasks, Transformers have shown favorable results compared to traditional ML and deep learning techniques.
Notably, Meng et al. \cite{meng2019spacecraft} applied the model to NASA telemetry datasets and achieved strong accuracy (0.78 F1) in detecting anomalies.
TranAD \cite{tuli_tranad_2022} proposed an adversarial training procedure to exaggerate reconstruction errors in anomalies.  Xu et al. \cite{xu_anomaly_2022} achieve state-of-the-art results in detecting anomalies in multivariate time series via association discrepancy.  Their key finding is that anomalies have high association with adjacent time points and low associations with the whole series, accentuating anomalies. 

Finally, Transformer variations have been proposed for time series forecasting to lower the attention complexity of long sequence time series \cite{li2019enhancing,zhou2021informer,liu2021pyraformer,pmlr-v162-zhou22g}, add stochasticity \cite{wu2020adversarial}, and incorporate traditional time series learning methods \cite{pmlr-v162-zhou22g,wu2021autoformer}.  Li et al. \cite{li2019enhancing} introduce LogSparse attention, which allows each cell to attend only to itself and its previous cells with an exponential step size.  The Informer method \cite{zhou2021informer} selects dominant queries to use in the attention module based on a sparsity measurement.  Pyraformer \cite{liu2021pyraformer} introduces a pyramidal attention mechanism for long-range time series, allowing for linear time and memory complexity.  Wu et al. \cite{wu2020adversarial} use a Sparse Transformer as a generator in an encoder-decoder architecture for time series forecasting, using a discriminator to improve the prediction.  

\subsection{Compressed Neural Networks}
Pruning unimportant weights from neural networks was first shown to be effective by Lecun et al. \cite{lecun_optimal_1989}.  In recent years, deep learning has scaled the size and computational cost of neural networks. Naturally, research has been directed at decreasing size \cite{han_learning_2015} and energy consumption \cite{yang_designing_2017} of deep learning models. 

The Lottery Ticket Hypothesis  \cite{frankle_lottery_2019} showed that randomly initialized neural networks contain sparse subnetworks that, when trained in isolation, achieve comparable accuracy to a trained dense network of the same structure.
The implications of this finding are that over-parameterized neural networks are no longer necessary, and we can prune large models and still maintain the original accuracy.

Subsequent work found that we do not need to train neural networks at all to find accurate sparse subnetworks; instead, we can find a high performance subnetwork using the randomly initialized weights \cite{ramanujan_whats_2020,malach_proving_2020,chijiwa_pruning_2021,gorbett2023randomly}.
Edge-Popup \cite{ramanujan_whats_2020} applied a scoring parameter to learn the importance of each weight, using the straight-through estimator \cite{bengio_estimating_2013} to find a high accuracy mask over randomly initialized models.  Diffenderfer and Kailkhuram \cite{diffenderfer_multi-prize_2021} introduced the \textit{Multi-Prize} Lottery Ticket Hypothesis, showing that 1) multiple accurate subnetworks exist within randomly initialized neural networks, and 2) these subnetworks are robust to quantization, such as binarization of weights.
In this work, we use the Biprop algorithm proposed in \cite{diffenderfer_multi-prize_2021} to binarize the weights of Transformer models.  

\subsection{Compressed and Efficient Transformers}
Large-scale Transformers such as the BERT (110 million parameters) are a natural candidate for pruning and model compression \cite{efficient_transformers,ganesh2021compressing}.
Chen et al. \cite{chen_pre-trained_2021} first showed that the Lottery Ticket Hypothesis holds for BERT Networks, finding accurate subnetworks between 40\% and 90\% sparsity.  
Jaszczur et al. \cite{jaszczur2021sparse} proposed scaling Transformers by using sparse variants for all layers in the Transformer.
Other works have reported similar findings \cite{lepikhin2020gshard,fedus2021switch}, showing that sparsity can help scale Transformer models to even larger levels.  
 
Other works have proposed modifications for more efficient Transformers aside from pruning \cite{efficient_transformers}.  Most research has focused on improving the $\mathcal{O}(n^2)$ complexity of attention, via methods such as fixed patterns \cite{qiu2019blockwise}, learnable patterns \cite{kitaev2020reformer}, low rank/kernel methods \cite{wang2020linformer,choromanski2020rethinking}, and downsampling \cite{zhang2021poolingformer,beltagy2020longformer}.  Various other methods have been proposed for compressing BERT networks such as pruning via post-training mask searches \cite{kwon2022fast}, block pruning \cite{lagunas2021block}, and 8-bit quantization \cite{zafrir2019q8bert}.  We refer readers to Tay et al. \cite{efficient_transformers} for details.

Despite the various works compressing Transformers, we were not able to find any research using both pruning and binarization.   Utilizing both methods allows for more efficient computation (measured using FLOPs) as well as a significant decrease in storage (due to binary weights).  Additionally, we find that our proposed model is still a fraction of the size of compressed NLP Transformers models when trained on time series tasks.  For instance, TinyBERT \cite{jiao-etal-2020-tinybert}  contains 14.5 million parameters and 1.2 billion FLOPs, compared to our models  which contain less than 1.5 million \textbf{binary} parameters and 38 million FLOPs.

\section{Method}\label{method}

Our model consists of a Transformer encoder \cite{vaswani_attention_2017} with several modifications.  
We base our model off of Zerveas et al. \cite{zerveas_transformer}, who propose using a common Transformer framework for several time series modeling tasks.  
To begin, we describe the base architecture of the Transformer as applied to multivariate time series.  Subsequently, we describe the techniques used for pruning and binarization.  Finally, we describe the two changes applied to the attention mechanism.

\subsection{Dense Transformer} 

We denote fully trained Transformers with no pruning and \gls{fp32} weights as Dense Transformers.  
Let $\mathbf{X_t} \in\mathbb{R}^{w\times m}$ be a model input for time $t$ with window size $w$ and $m$ features.  Each input contains $w$ feature vectors  $\mathbf{x} \in\mathbb{R}^{m}:\mathbf{X_t} \in\mathbb{R}^{w\times m}=[\mathbf{x_{t-w}},\mathbf{x_{t-w+1}},...,\mathbf{x_{t}} ]$, ordered in time sequence of size $w$.  In classification datasets $w$ is predefined at the sample or dataset level. For anomaly detection and forecasting tasks, we fix $w$ to 50 or 200 and use an overlapping sliding window as inputs.  
 
 The standard architecture (pre-binarization) projects $m$ features  onto a $d$-dimensional vector space using a linear module with learnable weights $\mathbf{W_p}\in\mathbb{R}^{d \times m}$
and bias $\mathbf{b_p}\in\mathbb{R}^{d}$.  
We use the standard positional encoder proposed by Vaswani et al.  \cite{vaswani_attention_2017}, and we refer readers to the original work for details.  For the Dense Transformer classification models, we use learnable positional encoder  \cite{zerveas_transformer}.  Zerveas et al. \cite{zerveas_transformer} propose using batch normalization instead of layer normalization used in traditional Transformer NLP models.  They argue that batch normalization mitigates the effects of outliers in time series data.  We found that for classification tasks, batch normalization performed the best, while in forecasting tasks layer normalization worked better.  For anomaly detection tasks we found that neither normalization technique was needed.  

Each Transformer encoder layer consists of a multi-head attention module followed by ReLU layers.   
The self-attention module takes input $\mathbf{Z_t}\in\mathbb{R}^{w \times d}$ and projects it onto a Query ($\mathbf{Q}$), Key ($\mathbf{K}$), and Value ($\mathbf{V}$), each with learnable weights $\mathbf{W}\in\mathbb{R}^{d \times d}$
and bias $\mathbf{b}\in\mathbb{R}^{d}$.  Attention is defined as
$
    Attention(\mathbf{Q}, \mathbf{K}, \mathbf{V}) =
Softmax\left(\frac{\mathbf{Q}\mathbf{K}^\intercal}{\sqrt{d}} \right) \mathbf{V}
$.  Queries, keys, and values are projected by the number of heads ($h$) to create multi-head attention. The resultant output $\mathbf{Z_t}'$ undergoes a nonlinearity before being passed to the next encoder layer.  The Transformer consists of $N$ encoder layers followed by a final decoder layer.  For classification tasks, the decoder outputs $l$ classification labels: $\mathbf{X'_t}\in\mathbb{R}^{w \times l}$, which are averaged over $w$. For anomaly detection and forecasting, the decoder reconstructs the full input: $\mathbf{X'_t}\in\mathbb{R}^{w \times m}$.  

\subsection{Sparse Binary Transformer}

Central to our binarization architecture is the Biprop algorithm \cite{diffenderfer_multi-prize_2021}, which uses randomly initialized floating point weights to find a binary mask over each layer.  
Given a neural network with weight matrix $\mathbf{W}\in\mathbb{R}^{i \times j}$ initialized with a standard method such as Kaiming Normal  \cite{he2015delving}, 
we can express a subnetwork over neural network $f(x;\mathbf{W})$ as $f(x;\mathbf{W}   \odot \mathbf{M})$, where $\mathbf{M}\in\{0,1\}$ is a binary mask and $\odot$ is an elementwise multiplication. 


To find $\mathbf{M}$, parameter $\mathbf{S}\in\mathbb{R}^{i \times j}$  is initialized for each corresponding $\mathbf{W}\in\mathbb{R}^{i \times j}$.  $\mathbf{S}$ acts as a score assigned to each weight dictating the importance of the weights contribution to a successful subnetwork.  Using backpropagation as well as the straight-through estimator \cite{bengio_estimating_2013}, the algorithm takes pruning rate hyperparameter $p \in [0,1]$, and on the forward pass computes $\mathbf{M}_k$ at layer $m$ as

\begin{equation}
    \mathbf{M}_k=   \begin{cases}
    1 & \text{if $|\mathrm{S}_k| \in \{\tau(k)_{k=1}^{l_m}\ge[l_m p]\}$}\\
    0 & \text{otherwise}
  \end{cases}
\end{equation}

where $\tau$ sorts indices $\{k\}^l_{k=1}\in\mathrm{S}$ such that $|S_{\tau(k)}|\le|S_{\tau(k+1)}|$.
Masks are computed by taking the absolute value of scores for each layer, and setting the mask to 1 if the value falls above the top $p^{th}$ percentile.   

To convert each layer to binary weights Biprop introduces gain term $\alpha\in\mathbb{R}$, which is common to \gls{bnns} \cite{qin2020binary}. The gain term utilizes floating-point weights \textit{prior} to binarization during training.  During test-time, the alpha parameter scales the binarized weight vector.  The parameter rescales binary weights $\mathbf{B} \in\{-1,1\}$ to $\{-\alpha,\alpha\}$, and the network function becomes $f(x;\alpha(\mathbf{B} \odot \mathbf{M}))$.  $\alpha$ is calculated as
\begin{equation}
 \alpha=\frac{||\mathbf{M}\odot\mathbf{W}||_1}{||\mathbf{M}||_1}
\end{equation}

with  $\mathbf{M}$ being multiplied by $\alpha$ for gradient descent (the straight-through estimator is still used for backpropagation). This calculation was originally derived by Rastegari et al.  \cite{rastegari2016xnor}.

In our approach we create sparse and binary modules for each linear and layer normalization layer.  Our model consists of two linear layers at the top most level: one for projecting the initial input (embedding in NLP models) and one used for the decoder output.  Additionally, each encoder layer consists of six linear layers: $\mathbf{Q}$, $\mathbf{K}$, and $\mathbf{V}$ projections, the multi-head attention output projection, and two additional layers to complement multi-head attention.

\subsection{Attention Modifications}\label{attentionmods}

In this section we describe two modifications made to the attention module
to reduce its quadratic complexity.  Several previous works have proposed changes to attention in order to lessen this bottleneck, such as Sparse Transformers \cite{child2019generating}, ProbSparse Attention \cite{zhou2021informer}, and Pyramidal Attention \cite{liu2021pyraformer}. While each of these works present quality enhancements to the memory bottleneck of attention, we instead seek to evaluate whether simple sparsification approaches can retain the accuracy of the model compared to canonical attention.  Our primary motivation for the following attention modifications are to test whether a compressed Transformer can retain the same accuracy as a Dense Transformer. 

\subsubsection{Fixed Q,K, and V Projection Mask}
To reduce the computational complexity of the matrix multiplications within the attention module, we apply random fixed masks to the  $\mathbf{Q}$, $\mathbf{K}$, and $\mathbf{V}$ projections.  We hypothesize that we can retain the accuracy of full attention by using this ``naive'' activation pruning approach, which requires no domain knowledge.  We argue that the success of this approach provides insight into the necessity of full attention computations. In other words, Transformers are expressive and powerful enough for certain tasks that we can prune the models in an unsophisticated way and maintain accuracy. Moreover, many time series datasets and datasets generated at the edge are often times simplistic enough that we can apply this unsophisticated pruning \cite{gorbett2022local,gorbett2022wip}.

To apply this pruning, on model initialization we create random masks with prune rate $p_a \in \{0,1\}$ for each attention module  and each projection Q,K, and V.  Attention heads within the same module inherit identical Q, K, or V masks.  The mask is applied to each projection during train and test. In each of our models we set the prune rate $p_a$ of the attention module equal to the prune rate of the linear modules ($p_a=p$).   

\subsubsection{Step-t Attention Mask}
For anomaly detection and single-step forecasting tasks, the \gls{sbt} algorithm relies on reconstructing or predicting outputs at the current time step $t$ for each feature $m$, despite $w$ time steps of data being provided to the model. Specifically, the \gls{sbt} model is only interested in input vector  $\mathbf{x_t}\in\mathbb{R}^{m}$. 
For anomaly detection, the model reconstructs $\mathbf{x_t}$ from the input, while in forecasting tasks the model masks $\mathbf{x_t}=0$ prior to model input, reconstructs the actual values during training and inference.  

\begin{figure}
\includegraphics[width=8cm]{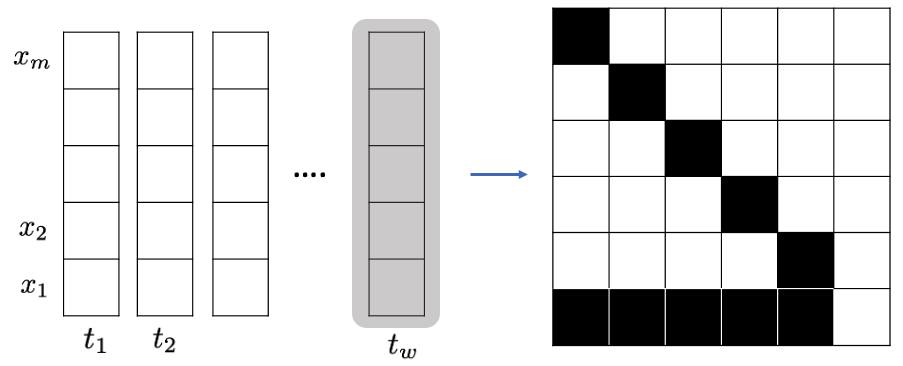}
\caption{\textbf{ Step-t Attention Mask}  Left: For the forecasting task we mask inputs during training in order to simulate unknown future time points. Right:  The Step-T attention mask used to calculate attention only at the current time-step versus past values.  Using this mask rather than setting our Query dimension to one enables us to pass time window vectors along multiple encoder layers.  }\label{steptfig}
\end{figure}

In both tasks, vector $\mathbf{x_t}$ contains the only values necessary for the model to learn, and our loss function reflects this by only computing error for these values.
As a result, computing attention for each other time step  adds unnecessary computation.  As depicted in Figure \ref{steptfig}, we pass a static mask to the attention module to compute attention only at step-T.  We additionally exclude attention computation at step-T with itself, forcing the variable to attend to historical time points for prediction.  Finally, we add diagonal ones to the attention mask at all past time points to add stability to training.  This masking method allows us to propagate the full input sample to multiple attention layers, helping us retain relevant historical information for downstream layers that would not be possible by changing the sizes of Q, K, and V to only model the $t$ time step.

\section{Experiments}


In this section we detail our experiments for time series classification, anomaly detection, and forecasting.  
Common to each learning task, we normalize each dataset prior to training such that each feature dimension $m$ has zero mean and unit variance.  We use the Transformer Encoder as described in Section \ref{method}, training each learning task and dataset  using the Dense Transformer and the \gls{sbt} to compare accuracy.  Finally, we run each experiment three times with a different weight seed, and present the average result.  For the \gls{sbt} model, varying the weight seed shows evidence of the robustness to hyperparameters.   Specific modifications to the model are made for each learning task, which we describe in the following sections. Additional training and architecture details can be found in the Appendix.

\subsection{Classification}

For our first time series learning task we select several datasets from the UCR Time Series Classification Repository \cite{bagnall16bakeoff}.  The datasets contain diverse characteristics including varying training set size (204-30,000), number of features (13-200), and window size (30-405).  We choose three datasets with the largest test set size (Insect Wingbeats, Spoken Arabic Digits, and Face Detection) as well as two smaller datasets (JapaneseVowels, Heartbeat).
Each dataset contains a set window size except for Insect Wingbeats and Japanese Vowels, which contain a window size \textit{up to} 30 and 29, respectively.  In these datasets, we pad samples with smaller windows to give them consistent window sizes. 
The decoder in our classification architecture is a classification head, rather than a full reconstruction of the input as is used in anomaly detection and forecasting tasks.  
The \gls{sbt} classification model is trained and tested using the fixed Q,K,V projection mask as described in Section \ref{attentionmods}.  

\noindent\textbf{Results}
In Table \ref{class_data}, we show that \gls{sbt}s perform as well as, or similar to, the Dense Transformer for each dataset at $p=0.5$ and $p=0.75$.  Our models are averaged over three runs with different weight seeds.   When comparing our model to state-of-the-art approaches, we find that the \gls{sbt} achieves strong results across each dataset, with the highest reported performance on three out of the five datasets.  Further, the \gls{sbt} models perform consistently across datasets while models such as Rocket \cite{dempster2020rocket} and Fran et al. \cite{franceschi2019unsupervised} have lower performance on one or more datasets.   

Surprisingly, the \gls{sbt} model achieves stronger average accuracy than the Dense Transformer (80.2\% versus 78.8\%), indicating that the pruned and binarized Transformer achieves a robust performance across datasets.  Despite this, Insect Wingbeats and Japanese Vowels datasets achieved a slightly lower performance at $p=0.5$ with a more substantial dropoff at $p=0.75$, indicating the model may lose some of its power on certain tasks.  



\begin{table}
\begin{center}
\renewcommand{\arraystretch}{1.3}%
\begin{tabular}{ lcccccc } 
\toprule
Model & \thead{Arabic \\Digits} & \thead{Heart \\Beat} & \thead{Insect\\W.B.} &\thead{Japan.\\Vowels}& \thead{Face\\Detect.}&Mean  \\
\midrule
XGBoost	&	69.6	&	73.2	&	36.9	&	96.2	&	63.3	&	67.8	\\
LSTM	&31.9	&	72.2	&	17.6	&	79.7	&	57.7	&	51.8	\\
Rocket \cite{dempster2020rocket}	&71.2	&	75.6	&	-	&	86.5	&	64.7	&	74.5	\\
Fran et al.	\cite{franceschi2019unsupervised}&95.6	&	75.6	&	16.0	&	98.9	&	52.8	&	67.8	\\
DTW\_D 	&	96.3	&	71.7	&	-	&	94.9	&	52.9	&	79.0	\\
Dense Trans.& 98.0	&	76.6	&	63.4	&	\textbf{98.0}	&	56.0	&	78.8	\\
\hdashline
$SBT_{p=0.5}$	&	98.2	&	77.2	&	\textbf{64.1}	&	95.3	&	\textbf{66.1}	&	\textbf{80.2}	\\
$SBT_{p=0.75}$	&\textbf{98.6}	&	\textbf{78.5}	&	61.3	&	85.3	&	65.8	&	77.9	\\
\bottomrule
\end{tabular}
\caption{\textbf{Accuracy of time series classification models on five datasets. }  Results are obtained from \cite{zerveas_transformer, franceschi2019unsupervised}.   \gls{sbt} models achieve higher accuracy than prior works (excluding the Dense Transformer) in each case, except for the Japanese Vowels dataset.  Additionally, \gls{sbt} models achieve accuracy within 2.7\% of the Dense Transformer for each dataset.        }\label{class_data}
\end{center}
\end{table}

\subsection{Anomaly Detection}

For the anomaly detection task we test the \gls{sbt} algorithm on established multivariate time series anomaly detection datasets used in previous literature: Soil Moisture Active Passive Satellite (SMAP) \cite{hundman2018detecting}, Mars Science Labratory rover (MSL) \cite{hundman2018detecting}, and the Server Machine Dataset (SMD) \cite{su_robust_2019}.  SMAP and MSL contain telemetry data such as radiation and temperature, while SMD logs computer server data such as CPU load and memory usage.  The datasets contain benign samples in the training set, while the test set contains labeled anomalies (either sequences of anomalies or single point anomalies). 


Our model takes sliding window data as input and reconstructs data at $\mathbf{x_t}$ given previous time points.  We use MSE to reconstruct each feature in $\mathbf{x_t}$. We use the step-T attention mask as described in Section \ref{method}.  
To evaluate our results, we adopt an adjustment strategy similar to previous works \cite{xu_anomaly_2022, su_robust_2019,shen2020timeseries,tuli_tranad_2022}: if \textit{any} anomaly is detected within a successive abnormal segment of time, we consider all anomalies in this segment to have been detected.  The justification is that detecting any anomaly in a time segment will cause an alert in real-world applications.  

To flag anomalies, we retrieve  reconstruction loss $\mathbf{x'_t}$ and threshold $\tau$, and consider anomalies where 
$\mathbf{x'_t}>\tau$.  Since our model is trained with benign samples, anomalous samples in the test set should yield a higher $\mathbf{x'_t}$. We compute $\tau$ using two methods from previous works:  A manual threshold \cite{xu_anomaly_2022} and the \gls{pot} method \cite{siffer2017anomaly}.  For the manual threshold, we consider proportion $r$ of the validation set as anomalous.  For SMD $r=0.5\%$, and for MSL and SMAP $r=1\%$.  For the \gls{pot} method, similar to OmniAnomaly \cite{su_robust_2019} and TranAd \cite{tuli_tranad_2022}, we use the automatic threshold selector to find $\tau$.  Specifically, given our training and validation set reconstruction losses, we use \gls{pot} to fit the tail portion of a probability distribution using the generalized Pareto Distribution.  \gls{pot} is advantageous when little information is known about a scenario, such as in datasets with an unknown number of anomalies.


\begin{table}
\begin{center}
\renewcommand{\arraystretch}{1.2}%
\begin{tabular}{ lccccc } 
\multirow{ 2}{*}{\small{Dataset} }  &\multirow{ 2}{*}{\small{Metric}} &\multicolumn{2}{c}{\underline{\small{Manual Threshold}} } &\multicolumn{2}{c}{ \small{\underline{\gls{pot} Threshold}} } \\
&& \small{ Dense } &$\gls{sbt}_{p=0.75}$& \small{ Dense } &$\gls{sbt}_{p=0.75}$\\
\midrule

\multirow{ 3}{*}{MSL} &P& 92.7&96.8&85.5&82.9\\ 
 &R& 100&100&100&100\\ 
 &F1&96.2&96.8&92.1&91.1\\ 
\midrule
\multirow{ 3}{*}{SMD} &P&85.4&85.3&99.9&100\\ 
 &R&100&100&100&100\\ 
 &F1&92.1&92.1&100&99.9\\  
\midrule
\multirow{ 3}{*}{SMAP} &P&93.9&93.7&85.9&84.9\\ 
 &R& 100&100&100&100\\ 
 &F1& 96.9&96.8&92.4&91.8\\ 
\bottomrule
\end{tabular}
\caption{Anomaly detection results with benign sample windows.  We evaluate  Precision (P), Recall (R), and the F1 score using both manual threshold and \gls{pot} threshold technique.  We find that the single time step prediction window achieves high accuracy when each past time-step in $w$ is benign.  $w=200$ for SMD and $w=50$ for SMAP and MSL. These results indicate that when given time to stabilize after an anomalous event, our \gls{sbt} framework can detect new anomalies with high accuracy.  We evaluate our results using a manual threshold ($\tau$=0.5\% for SMD, 1\% for others) and the \gls{pot} automatic threshold selector. }\label{anomaly_data}
\end{center}
\end{table}

\noindent\textbf{Results}
In Table \ref{anomaly_data} we report the unique findings of our single-step anomaly detection method using Precision, Recall, and F1-scores.  Specifically, we find that when only considering inputs with fully benign examples in window $w$, both the \gls{sbt} and the Dense Transformer achieve high accuracy on all three datasets (F1 between 90.6 and 100).  In other words, we find that our model performance is best when we filter examples that have an anomalous sequence or data point in $[\mathbf{x_{t-w}}, \mathbf{x_{t-w+1}},...,\mathbf{x_{t-1}}]$. For SMD, $w=200$ and  for SMAP and MSL $w=50$.  This observation implies that the model needs time to stabilize after an anomalous period.  Intuitively, if an anomaly occurred recently, new benign observations will have a higher reconstruction loss as a result of their difference with the anomalous examples in their input window.  We argue that this validation metric is logical in real-world scenarios, where monitoring of a system after an anomalous period of time is necessary.  

We additionally report F1-scores compared to state-of-the-art time series anomaly detection models in Table \ref{anomaly_sota}.  To accurately compare our model against existing methods, we use the full test set without filtering out benign inputs with anomalies in the near past.  \gls{sbt} results are much more modest, with F1-scores between 70 and 88.  Despite this, our method still performs stronger than non-temporal algorithms such as the Isolation Forest, as well as other deep-learning based approaches such as Deep-SVDD and BeatGan.  





\begin{table}
\begin{center}
\renewcommand{\arraystretch}{1.2}%
\begin{tabular}{ lccc | c } 
Model  & SMD & MSL& SMAP& Avg. \\
\midrule
LOF	&	46.7	&	61.2	&	57.6	&	55.2	\\
IsolationForest	&	53.6	&	66.5	&	55.5	&	58.5	\\
OCSVM	&	56.2	&	70.8	&	56.3	&	61.1	\\
DAGMM	&	57.3	&	74.6	&	68.5	&	66.8	\\
VAR	&	74.1	&	77.9	&	64.8	&	72.3	\\
MMPCACD	&	75.0	&	70.0	&	81.7	&	75.6	\\
ITAD	&	79.5	&	76.1	&	73.9	&	76.5	\\
Deep-SVDD	&	79.1	&	83.6	&	69.0	&	77.2	\\
$\textbf{\gls{sbt}}_{p=0.9}$	&	82.5	&	78.5	&	70.6	&	77.2	\\
CL-MPPCA	&	79.1	&	80.4	&	72.9	&	77.5	\\
BeatGAN	&	78.1	&	87.5	&	69.6	&	78.4	\\
$\textbf{\gls{sbt}}_{p=0.5}$	&	87	&	78.4	&	69.8	&	78.4	\\
$\textbf{\gls{sbt}}_{p=0.75}$	&	88.0	&	79.3	&	70.6	&	79.3	\\
LSTM-VAE	&	82.3	&	82.6	&	78.1	&	81.0	\\
OmniAnomaly	&	85.2	&	87.7	&	86.9	&	86.6	\\
Anomaly Transformer	&	92.3	&	93.6	&	96.7	&	94.2	\\
\bottomrule
\end{tabular}
\caption{\textbf{F1 scores of various time series anomaly detection models. } We compare our \gls{sbt} framework with several state-of-the-art algorithms on the anomaly detection task.  The table is ordered by average F1 accuracy across each dataset.  We evaluate our algorithm using the traditional method (different from Table \ref{anomaly_data}), where each sample can contain anomalous events in its input window.  We use a manual threshold to report results for the \gls{sbt} model.}\label{anomaly_sota}
\end{center}
\end{table}

\begin{table}[b]
\begin{center}
\renewcommand{\arraystretch}{1.45}%
\begin{tabular}{ lcccccc } 
\toprule
\multirow{ 2}{*}{Model} & \multicolumn{2}{c}{ECL}& \multicolumn{2}{c}{Weather}& \multicolumn{2}{c}{ETTm1}  \\
&MSE&MAE&MSE&MAE&MSE&MAE\\
\midrule
Informer& 0.185&0.301&0.159&0.197&0.051&0.150\\
Pyraformer & 0.149&0.305&-&-&0.081&0.214\\
Dense Trans.& 0.182&0.299&0.173&0.225&0.070&0.201\\
\hdashline
$\gls{sbt}_{p=0.5}$& 0.198&0.316&0.166&0.216&0.059&0.171\\
$\gls{sbt}_{p=0.75}$& 0.221&0.333&0.168&0.218&0.070&0.191\\
\bottomrule
\end{tabular}
\caption{A summary of time series forecasting models on three datasets.  Each \gls{sbt} model is run three times with different weight seeds and averaged.  Standard deviation is less than 0.01.    }\label{forecasting_results}
\end{center}
\end{table}

\subsection{Forecasting}

We test our method on single-step forecasting using the Step-T attention mask. Specifically, using the framework outlined by Zerveas et al. \cite{zerveas_transformer}, we train our model by masking the input at the forecasting time-step $t$.  For example, input
$\mathbf{X_t}$ containing $m$ features and $t$ time-steps
$[\mathbf{x_{t-w}},\mathbf{x_{t-w+1}},...,\mathbf{x_{t}} ]$ is passed through the network with 
 $\mathbf{x_{t}}=0$.  We then reconstruct this masked input with the Transformer model, using mean squared error between the masked inputs reconstruction and the actual value.  The masking method simulates unseen future data points during train time, making it compatible with the forecasting task during deployment.

We test our model on three datasets used in previous works:  ECL contains electricity consumption of 321 clients in Kwh.  The dataset is converted to hourly consumption values due to missing data. Weather contains data for twelve hourly climate features for 1,600 location in the U.S.  ETTm1 (Electricity Transformer Temperature) contains 15-minute interval data including oil temperature and six additional power load features.  Additional training details are available in the Appendix. 

We compare our method against the Informer \cite{zhou2021informer} and the Pyra-former \cite{liu2021pyraformer}  trained with single-step forecasting.  Both are current state-of-the-art models that have shown robust results compared against a variety of forecasting techniques.  Importantly, each method is compatible with multivariate time series forecasting as opposed to some research.  We note that these models are built primarily for \gls{lstf}, which we do not cover in this work.  

\begin{figure}[t]
\includegraphics[width=8.5cm]{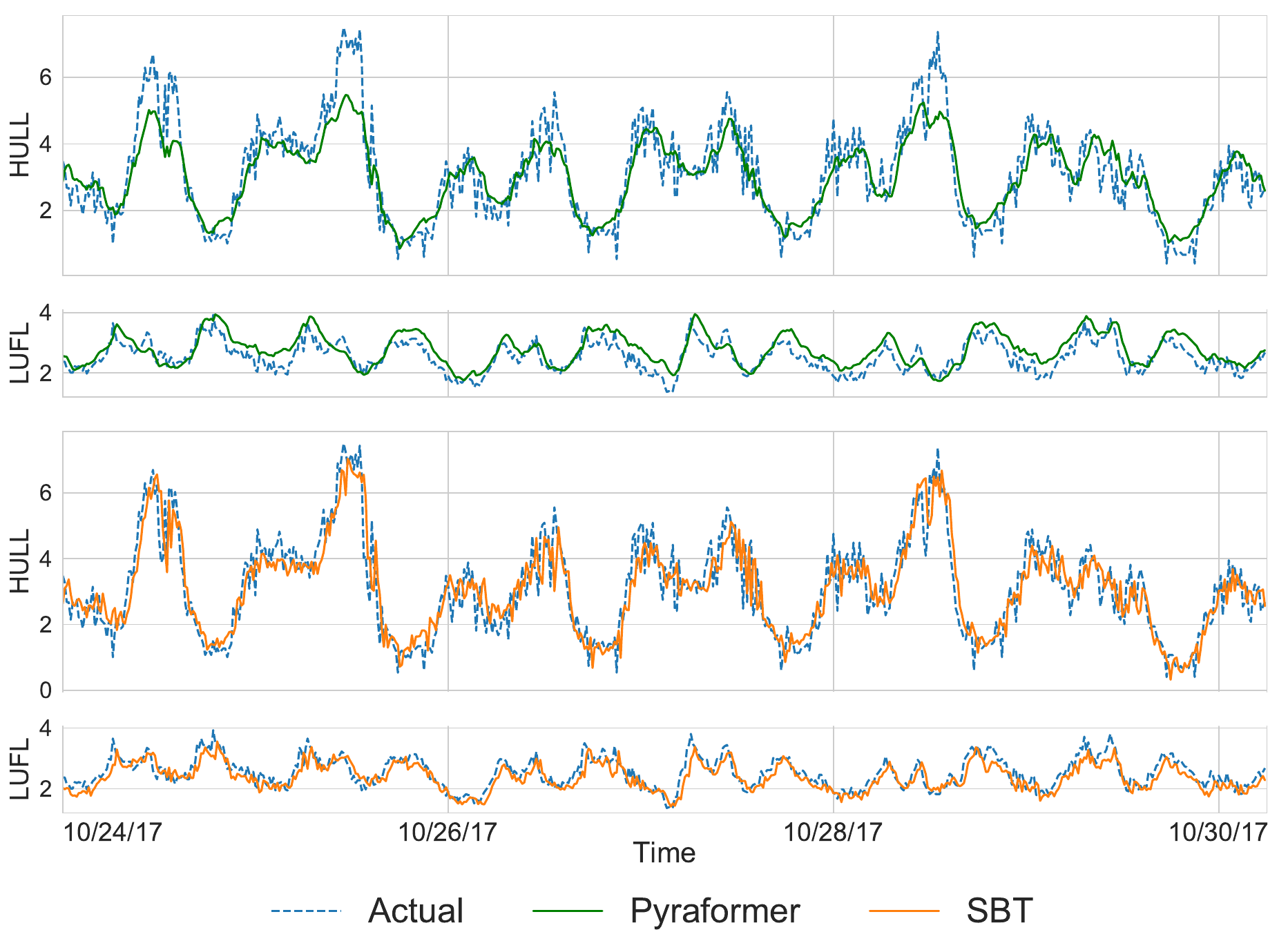}
\caption{Time series predictions on the ETTm1 dataset for the Pyraformer (top) and Sparse Binary Transformer (bottom) .  We show 600 predictions across each model for two features (HULL, LUFL).    }\label{forecast_pic}
\end{figure}

\noindent\textbf{Results}
We evaluate results in Table \ref{forecasting_results} using MSE and MAE on the test set of each dataset.  Results indicate that the \gls{sbt} model achieves accuracy comparable to the Dense architecture in each dataset at $p=0.5$.  Interestingly, the Weather at ETTm1 \gls{sbt} models achieved \textit{better} accuracy than the dense model at $p=0.5$.  Both models additionally showed robustness to higher prune rates, with accuracy dropping off slowly.  ECL on the other hand showed some sensitivity to prune rate, with a slight drop off when increasing the prune rate.  We find that datasets with a higher dimensionality performed the worst: ECL contains 321 features, while Insect Wingbeats contains 200.  Increasing the dimensionality of the model ($d$) mitigated some of these effects, however it was at the cost of model size and complexity.  Despite this, we find that the \gls{sbt} model is able to predict the general trend of complex patterns in data, as depicted in Figure \ref{forecast_pic}.  

Compared to state-of-the-art approaches such as the Pyraformer and Informer architectures, our general purpose forecasting approach performs comparably, or slightly worse, on the single-step forecasting task.  Metrics were not substantially different for any of the models except for the ECL dataset, where Pyraformer was easily the best model.  Comparing the architectures, we find that the \gls{sbt} model achieves substantially lower computational cost than both the Informer and Pyraformer models.  For example, on the ECL dataset, Pyraformer contains 4.7 million parameters and the Informer 12.7 million parameters (both \gls{fp32}, while the \gls{sbt} model contains 1.5 million binary parameters.

\subsection{Architecture}
Each model in our framework consists of 2 encoder layers each with a multi-head attention module containing two heads.  The feedforward dimensionality for each model is 256 with ReLU used for nonlinearity.  Classification models had the best results using Batch Normalization layers, similar to \cite{zerveas_transformer}, while forecasting models used Layer Normalization typical of other Transformer models.  For anomaly detection we did not use Batch or Layer Normalization.  For the output of our models, 
anomaly detection and forecasting rely on a single decoder linear layer which reconstructs the output to size ($m$, $w$), while classification outputs size ($d$, $num. classes$) and takes the mean of $d$ to formulate a final classification prediction.  Further details are included in the Appendix and the code repository.

\section{Computational Savings}

In this section we estimate the computational savings achieved by using  the \gls{sbt} model.  
We will begin by introducing the metrics used to estimate computational savings, and will then summarize the results of these metrics for each model and task.

We note that several works (highlighted in Section \ref{related}) have proposed modifications to the Transformer in order to make attention more efficient.  In this section, we concentrate on the  enhancements achieved by 1) creating a sparsely connected Transformer with binary weights, and 2) simplifying the attention module for time series specific tasks such as single-step prediction and classification.  We argue that these enhancements are independent of the achievements made by previous works.

\subsection{Metrics}

\noindent \textbf{FLOPs (Non-zero).} In the field of network pruning, \gls{flops}, or the number of multiply-adds, is a commonly used metric to quantify the efficiency of a neural network \cite{blalock2020state}.  The metric computes the number of floating point operations required for an input to pass through a neural network.  We use the ShrinkBench tool to calculate FLOPs, a framework 
proposed by Blalock et al. \cite{blalock2020state} to perform standardized evaluation on pruned neural networks.

Our Transformer architecture contains \gls{fp32} activations at each layer along with binary weights scaled to $\{-\alpha,\alpha\}$.  As a result, no binary operations are performed, and our total FLOPs count is a function of prune rate $p$.  For example, a linear module with a standard FLOPs count of $d \times m$ has a new FLOPs count of $d \times m \times p$, where $p \in [0,1]$.  Linear layers outside of attention do not need window size added to the matrix multiply because the inputs are permuted such that batch size is the second dimension of the layer input.  Each equation counts the number of \textit{nonzero} multiply-adds necessary for the neural network.

\begin{table}[ht]
\begin{center}
\renewcommand{\arraystretch}{1.4}%
\begin{tabular}{ lccc } 
\toprule
Attention Type & Q,K,V Proj.& $\mathbf{Q}\mathbf{V}^\intercal$& $\mathbf{AV}$   \\
\midrule
Canonical&$d^2w$&$d'w^2$&$d'w^2$ \\ 
Step-T Mask&$(d^2w)p_a$&$(w-1)d'$&$2(w-1)d'$  \\
Q,K,V Mask &$(d^2w)p_a$&$d'(wp_a)^2$&$(d'w^2)p_a$ \\
\hline
\end{tabular}

\end{center}
\caption{\textbf{Non-zero FLOPs equations for various attention modules.}  These calculations assume $\mathbf{Q},\mathbf{K}$ and $\mathbf{V}$ are equal sized projections in $\mathbb{R}^{w \times d}$, and $d'=d/h$.  $\mathbf{Q}\mathbf{V}^\intercal$ and $\mathbf{AV}$ are additionally multiplied by $h$.  $\mathbf{Q}$-scaling and softmax FLOPs excluded from this table. }\label{attention_equations}
\end{table}

\begin{table*}
\large
\begin{center}
\setlength{\tabcolsep}{0.25em}
\renewcommand{\arraystretch}{1.45}%
\begin{tabular}{    c|lccc|ccc|cccc|cc } 
\toprule
\multicolumn{1}{c}{}&\multicolumn{4}{c}{}&\multicolumn{3}{c}{Dense Transformer}&\multicolumn{4}{c}{Sparse Binary Transformer}&\multicolumn{2}{c}{ $\sim$Savings}\\
\midrule
\multicolumn{1}{c}{Type}&\multicolumn{1}{c}{Dataset}&$m$&$w$&\multicolumn{1}{c}{$d$}&\thead{Params (\gls{fp32})\\ (K)}& \thead{Size (Bits)\\ (Mil.)}& \multicolumn{1}{c}{\thead{FLOPs  \\(Mil.)}} &$p$ & \thead{Params (Binary) \\(K)}& \thead{Size (Bits)\\ (Mil.)}&\multicolumn{1}{c}{\thead{FLOPs  \\(Mil.)}} &\thead{Size  (Bits) \\($\frac{\text{Dense}}{\text{\gls{sbt}}}$)}&\thead{FLOPs  \\($\frac{\text{Dense}}{\text{\gls{sbt}}}$)}\\
\hline

 \parbox[t]{2mm}{\multirow{5}{*}{\rotatebox[origin=c]{90}{\footnotesize{Classification}}}}&Heartbeat	&	61	&	405	&	64	&	169.6	&	5.4	&	52.7	&	0.5	&	102.3	&	0.1	&	21.6	&	$\times$49.1	&	$\times$2.4	\\
&Insect W.B.	&	200	&	30	&	128	&	555.5	&	17.8	&	5.4		&	0.5	&	420.1	&	0.4	&	2.7	&	$\times$40.0	&	$\times$2.0	\\
&Arabic Dig.	&	13	&	93	&	64	&	167.1	&	5.3	&	2.8	&	0.5	&	100.0	&	0.1	&	1.3	&	$\times$49.5	&	$\times$2.2	\\
&Japan.Vowels	&	12	&	29	&	32	&	75.5	&	2.4	&	0.3	&	0.5	&	41.6	&	0.04	&	0.2	&	$\times$52.9	&	$\times$2.1	\\
&FaceDetection	&	144	&	62	&	128	&	414.9	&	13.3	&	8.3	&	0.5	&	281.3	&	0.3	&	4.0	&	$\times$44.7	&	$\times$2.1	\\
\hdashline
 \parbox[t]{7.2mm}{\multirow{3}{*}{\rotatebox[origin=c]{90}{\footnotesize{\thead{Anomaly\\Detection}}}}}&MSL	&	55	&	50	&	110	&	223.7	&	7.2	&	4.9	&	0.75	&	221.5	&	0.2	&	1.0	&	$\times$32.3	&	$\times$5.0	\\
&SMAP	&	25	&	50	&	50	&	75.2	&	2.4	&	1.3	&	0.75	&	73.7	&	0.1	&	0.2	&	$\times$32.6	&	$\times$6.1	\\
&SMD	&	38	&	200	&	76	&	132.8	&	4.2	&	19.5	&	0.75	&	129.8	&	0.1	&	1.9	&	$\times$32.7	&	$\times$10.5	\\
\hdashline
 \parbox[t]{2mm}{\multirow{3}{*}{\rotatebox[origin=c]{90}{\footnotesize{Forecast.}}}}&ECL	&	321	&	200	&	350	&	1569.4	&	50.2	&	204.8	&	0.75	&	1563.9	&	1.6	&	74.5	&	$\times$32.1	&	$\times$2.7	\\
&Weather	&	7	&	200	&	100	&	188.0	&	6.0	&	28.5	&	0.5	&	185.6	&	0.2	&	6.2	&	$\times$32.4	&	$\times$4.6	\\
&ETTm1	&	12	&	200	&	64	&	102.0	&	3.3	&	15.5	&	0.5	&	100.0	&	0.1	&	2.6	&	$\times$32.6	&	$\times$5.9	\\
\hline
\end{tabular}
\caption{\textbf{Computational savings for Dense Transformers compared to \glspl{sbt}}.  
 \gls{sbt} models achieve a substantial reduction in size and FLOPs count across all models.  We denote parameters in thousands and size and FLOPs in millions, with savings calculated by dividing the Dense values by the \gls{sbt} values.}\label{table:flops_results}

\end{center}
\end{table*}

Furthermore, we modify the FLOPs for the attention module to account for step-t attention mask and the fixed $\mathbf{Q}, \mathbf{K}, \mathbf{V}$ mask, as summarized in Table \ref{attention_equations}.  
In the standard attention module where $\mathbf{Q},\mathbf{K}$ and $\mathbf{V}$ are equal sized projections,  matrix multiply operations ($\mathbf{Q}\mathbf{V}^\intercal$, $\mathbf{AV}$) for each head equate to $d'w^2$, where $d'=d/h$.  
For step-t attention, we only require computation at the current time step (the last row in Figure \ref{steptfig}), while each each of the identities for past time steps equates to one.  $\mathbf{A} \mathbf{V}$ requires double the computations because $\mathbf{V}$ contains \gls{fp32} activations multiplied by the diagonal in $\mathbf{A}$.  For the fixed mask, 
since $\mathbf{Q}$ and $\mathbf{K}$ are sparse projections, we only require $(wp_a)^2$ nonzero computations in the matrix multiply.  Since  $\mathbf{A}$ is a dense matrix, we require $w^2$ FLOPs to multiply  sparse matrix $\mathbf{V}$. 

A simplified equation for network FLOPs becomes $2L+N(2L+MHA)$, where $L$ is a linear layer, $N$ is the number of attention layers, and $MHA$ is the multihead attention FLOPs (details described in Table \ref{attention_equations}).  Several FLOP counts are omitted from this equation, which we include in our code, including positional encoding, $Q$-scaling, and layer and batch norm. 



\noindent \textbf{Storage Size.} We measure the size of each model in total bits.  Standard networks rely on weights optimized with the \gls{fp32} data type (32 bits).  We consider each binarized module in our architecture to contain single bit weights with a single \gls{fp32} $\alpha$ parameter for each layer.  Anomaly detection and classification datasets contain 14 binarized modules, and forecasting contains 18 with the additional binarization of the layer normalization.   We note that the binarized quantities are only theoretical as a result of the PyTorch framework not supporting the binary data type.  Hardware limitations are also reported in other works \cite{frankle_lottery_2019}.

\subsection{Model Size Selection}\label{size}
Important to our work is tuning the size of each model.  We analyze whether we can create a Dense Transformer with a smaller number of parameters and still retain a performance on par with a larger model.  Our motivation for model size selection is two-fold: 1)  Previous research has found that neural networks need to be sufficiently overparameterized to be pruned and retain the same accuracy of the dense model and 2) The time series datasets studied in this paper have a smaller number of dimensions than the vision datasets studied in most pruning and model compression papers.  The effect of model overparameterization is that we need a dense model with enough initial parameters in order to prune it and still retain high performance.  Theoretical estimates on the number of required parameters are proposed by the Strong Lottery Ticket Hypothesis \cite{pensia_optimal_2020, orseau_logarithmic_2020} and are further explored in other pruning papers \cite{diffenderfer_multi-prize_2021,chijiwa_pruning_2021}.  On the other hand, the limited features of some time series datasets (such as Weather with 7 features) leads us to wonder whether we could simply create a smaller model.  

To alter the model size, we vary the embedding dimension $d$ of the model.  To find the ideal size of the model, we start from a small embedding dimension (such as 8 or 16), and increase the value in the Dense Transformer until the model performance on the validation set stops increasing.  With this value of $d$, we test the \gls{sbt} model.  

Our results show that in each dataset, Dense Transformers with a smaller embedding dimension $d$ either a) perform worse than the \gls{sbt} at the optimized size, b) contain more parameters (as measured in total bits), c) have more FLOPs, or d) some combination of the above.  In almost every dataset, the smaller Dense Transformer performs worse than the \gls{sbt} while also requiring more size and FLOPs.  The exception to this was Spoken Arabic Digits, where the smaller Dense Transformers ($d=16$ and $d=32$) performed slightly better than the \gls{sbt} with $d=64$.  Additionally, these models had a lower FLOPs count.  The advantage of the \gls{sbt} model in this scenario was a substantially lower storage cost than both smaller Dense models.  Even if both Dense Transformer models were able to be quantized to 8-bit weights, the storage of the \gls{sbt} would still be many times lower.  The ETTm1 dataset additionally had high performance Dense Transformers with a smaller size ($d=16, d=32)$.  However, both models were substantially more costly in terms of storage and additionally had a higher FLOPs count.  Detailed results are provided in the Appendix.

\subsection{Analysis} 

Results in Table \ref{table:flops_results} highlight the large computational savings achieved by \gls{sbt}.   
We find that layer pruning reduces FLOPs count (due to the added nonzero computations), while binarization helps with the storage size. 

Notably, all models have a FLOPs count at least two times less than the original Dense model. FLOPs are dramatically reduced in the anomaly detection and forecasting datasets, largely due to the step-t masking. 
Classification datasets have a dense attention matrix, leading to a smaller FLOPs reduction due to the softmax operation and the $AV$ calculation (where $V$ is sparse).  We note that using a higher prune rate can reduce the FLOPs more, however we include results at 50\% prune rate for classification since these models achieved slightly better accuracy.  

We highlight the storage savings of \gls{sbt} models by measuring bit size and parameter count.  Table \ref{table:flops_results} summarizes the substantial reduction in bit size for every model, with only two \gls{sbt} models having a bit size greater than 1 million (Insect Wingbeats and ECL). 
The two models with a larger size also had the highest dimensionality $m$, and consequently $d$.  
  
We note that  \gls{sbt} models contain a small number of \gls{fp32} values due to the single $\alpha$ parameter in each module.  Additionally, we forego a learnable encoding layer in \gls{sbt} classification models, leading to a smaller overall count.  Finally, no bias term is added to the \gls{sbt} modules, leading to a smaller number of overall parameters.  

Compared to other efficient models, our model generally has a lower FLOPs count. For example, MobileV2 \cite{sandler2018mobilenetv2} has 16.4 million FLOPs when modeling CIFAR10, while EfficientNetV2 \cite{tan2021efficientnetv2} has 18.1 million parameters.  

\section{Discussion}
We show that Sparse Binary Transformers attain similar accuracy to the Dense Transformer across three multivariate time series learning tasks: anomaly detection, forecasting, and classification.  We estimate the computational savings of \gls{sbt}'s by counting FLOPs as well as total size of the model.  

\subsection{Applications}
\gls{sbt}s retain high performance compared to dense models, coupled with a large reduction in computational cost. As a result, 
\gls{sbt}s have the potential to impact a variety of new domains.  
For example, sensors and small embedded systems such as IoT devices could employ  \gls{sbt}s for intelligent and data-driven decisions, such as detecting a malicious actor or forecasting a weather event.  
Such devices could be extended into new areas of research such as environmental monitoring.  Other small capacity applications include implantable devices, healthcare monitoring, and various industrial applications.  

Finally, lightweight deep learning models can also benefit larger endeavors. For example, space and satellite applications, such as in the MSL and SMAP telemetry datasets, collect massive amounts of data that is difficult to monitor.  Employing effective and intelligent algorithms such as the Transformer could help in the processing and auditing of such systems. 


\subsection{Limitations and Future Work}
Although \gls{sbt}s theoretically reduce computational costs, the method is not optimized for modern libraries and hardware.  Python libraries do not binarize weights to single bits, but 8-bit counts.  Special hardware in IoT devices and satellites could additionally make implementation a burden.  
Additionally, while our implementation shows that sparse binarized Transformers exist, the Biprop algorithm requires backpropagation over a dense network with randomly initialized \gls{fp32} weights.  Hence, finding accurate binary subnetworks requires more computational power during training than it does during deployment.  This may be a key limitation in devices seeking autonomy.
In addition to addressing these limitations, a logical step for future work would be to implement \gls{sbt}s in state-of-the-art Transformer models such as the Pyramformer for forecasting and the Anomaly Transformer for time series anomaly detection.    

\gls{sbt}s have the potential to enable widespread use of AI across new applications.  The Transformer stands as one of most powerful deep learning models in use today, and expanding this architecture into new domains provides promising directions for the future.

\section{Acknowledgements}
This work was supported in part by funding from NSF under Award
Numbers ATD 2123761, CNS 1822118, NIST, ARL, Statnett, AMI,
NewPush, and Cyber Risk Research.

\bibliographystyle{ACM-Reference-Format}
\balance
\bibliography{sbt}

\appendix
\setcounter{figure}{0}
\setcounter{table}{0}

\section*{Supplemental Materials}

\section{Ablation Studies}
We conduct two ablation studies testing the effects of removing the individual pruning mechanisms from the attention computation. We note that the attention pruning methods complement Biprop -- Biprop mainly reduces the model size, whereas attention pruning does a better job at  reducing the FLOPs. Each ablation experiment is averaged over three experimental runs with different seeds.  

Table \ref{table:ablation1} highlights the effects of removing random pruning from the time series classification models.  Notably, Biprop plus random pruning performs comparably to, or better than, Biprop on its own.  Adding random pruning even outperforms using only Biprop with the Japanese Vowels dataset.

Table \ref{table:ablation2} highlights the results of attention variations for both anomaly detection and forecasting tasks. Specifically, we look at our proposed approach (Biprop+Step-T Mask), Biprop plus an identity matrix mask in the attention layers, and finally Biprop only.   We report results using mean squared error (MSE) loss averaged over three runs.  

Results show that Biprop plus the Step-T mask performs comparably to using Biprop only.  For anomaly detection tasks, the MSE is even lower compared to just using Biprop.  Comparing both methods to the Biprop plus the identity matrix attention mask, we can see a significant difference in the results: the identity matrix attention mask attains a higher loss in each case.  
\begin{table}[h!]
\large
\begin{center}
\setlength{\tabcolsep}{0.25em}
\begin{tabular}{ ccc }
\hline
Dataset& Biprop+Random Pruning & Biprop  \\
\hline
Arabic Digits&	98.2&	98.2\\
Heartbeat	&77.7&	77.1\\
Insect Wingbeats	&64.1	&64\\
Japanese Vowels	&95.3&	84.4\\
Face Detection	&66.1&	65.9\\

\hline
\end{tabular}
\caption{We compare Biprop with Biprop plus random pruning on classification tasks.  We find that random pruning of the attention activations does not hurt classifcation accuracy, and in fact helps it in the case of the Japanese Vowels dataset. }\label{table:ablation1}

\end{center}
\end{table}

\begin{table}[h]
\large
\begin{center}
\setlength{\tabcolsep}{1.2em}
\begin{tabular}{ cccc }
\hline
Dataset&	\thead{Biprop+\\Step-T}	&\thead{Biprop+\\Identity Matrix}	& \thead{Biprop\\Only}\\
\hline
\multicolumn{4}{c}{Anomaly Detection}\\	
\hline
MSL	&0.277&	0.364	&0.357\\
SMAP&	0.117&	0.131&	0.125\\
SMD&	0.037&	0.041&	0.052\\
\hline
\multicolumn{4}{c}{Forecasting}\\	
\hline
ETTm1&	0.059&	0.068&	0.070\\
ECL&	0.198&	0.204&	0.182\\
Weather&	0.166&	0.180	&0.166 \\
\hline
\end{tabular}
\caption{We compare Biprop plus the Step-T attention mask with two other methods.  We find that Biprop with the Step-T mask performs similarly to using Biprop with full attention (Biprop Only).  Biprop with an Identity Mask on the attention computation performs worse than the other two methods.  We report results using MSE loss averaged across three runs.  }\label{table:ablation2}

\end{center}
\end{table}

\section{Training Details}

Each model is trained with Adam optimization with a learning rate of 1e-3 except for InsectWingbeats, where we use a learning rate of 1e-4.  For Dense Transformer classification models we use a learnable positional encoding, while in all other models we use a standard positional encoding.  

We found that \gls{sbt} models sometimes take slightly longer to converge, hence we train the models for more epochs in the forecasting and classification tasks. These numbers are specified in the configuration files in the code repository. 
Batch Normalization is used for classification tasks, layer normalization is used for forecasting tasks, and no normalization is used for anomaly detection.


\section{Analysis}
\subsection{Attention Magnitude Pruning versus Random Pruning}

As apart of our attention pruning analysis, we also applied magnitude pruning to the attention layers.  However, this method requires extra computation as a result of the sorting required to take the top activation's for each input.  Below we compare the results of magnitude pruning versus random pruning, finding that random pruning achieves similar accuracy to magnitude pruning at a lower computational cost. 

\begin{table}[h]
\large
\begin{center}
\setlength{\tabcolsep}{1.2em}
\begin{tabular}{ ccc }
\hline
Dataset&	 Mag. Prune	&Rand. Prune\\
	
\hline
Arabic 	&98.2	&98.2\\
Heartbeat	&77.2&	77.7\\
Insect 	&64.4&	64.1\\
Japanese &	94.9&	95.3\\
Face Det.	&66.3	&66.1\\

\hline
\end{tabular}
\caption{Random pruning versus activation magnitude pruning.  We find that random pruning achieves similar accuracy to magnitude pruning with lower computational cost.   }\label{table:other1}

\end{center}
\end{table}

\subsection{Model size savings of Biprop versus Pruning}

In Table \ref{table:other2} we compare the model size savings of Biprop compared to 32-bit pruning as well as pruning plus quantization (8-bit).  We show that, even compared to pruning plus 8-bit quantization, Biprop achieves substantially lower model size.

\begin{table}[h]
\large
\begin{center}
\setlength{\tabcolsep}{1.2em}
\begin{tabular}{ cccc }
\hline
Dataset&	Bin.&	Prune& \thead{Pruning+\\Quantization}	  \\
	
\hline
\multicolumn{4}{c}{Classification}\\	
\hline
Heartbeat&	$\times$ 49.10&	$\times$3.2&	$\times$11.7\\
Insect &	$\times$ 39.98	&$\times$ 2.6&	$\times$ 9.4\\
Arabic &	$\times$ 49.51&	$\times$ 3.2&	$\times$ 11.8\\
Japanese&	$\times$ 52.85&	$\times$ 3.5&	$\times$ 12.8\\
FaceDet.&	$\times$ 44.67&	$\times$ 2.8&	$\times$ 10.5\\
\hline
\multicolumn{4}{c}{Anomaly Detection}\\	
\hline
MSL&	$\times$ 32.31&	$\times$ 3.7&	$\times$ 11.8\\
SMAP&	$\times$ 32.65&	$\times$ 3.7&	$\times$ 11.9\\
SMD&	$\times$ 32.74&	$\times$ 3.7&	$\times$ 11.9\\
\hline
\multicolumn{4}{c}{Forecasting}\\	
\hline
Electricity&	$\times$ 32.11&	$\times$ 3.7&	$\times$ 11.7\\
Weather&	$\times$ 32.42&	$\times$ 2.0&	$\times$ 7.2\\
ETTm1&	$\times$ 32.65&	$\times$ 2.0&	$\times$ 7.3\\

\hline
\end{tabular}
\caption{ Comparison of the size between Biprop, 32-bit pruning, and 32-bit pruning + quantization.  Biprop achieves the greatest model size compression by a large degree.  }\label{table:other2}

\end{center}
\end{table}

\section{Dataset Details}
We report the details of datasets used for each task below.  For anomaly detection and forecasting tasks, we set the window size $w$ to a fixed value, while in classification, $w$ is predefined.  



\begin{table}[h]
\begin{center}
\begin{tabular}{ |c|c|c|c|c|c| } 
\hline
Dataset & Train Size & Test Size & $m$ & $w$ & Classes \\
\hline
Arabic& 6,599&2,199&13&93&10 \\ 
Heartbeat& 204&205&61&405&2 \\ 
Insect& 30k&20k&200&30&10 \\ 
Japanese & 270&370&12&29&9 \\ 
Face Det.& 5,890&3,524&144&62&2 \\ 
\hline
\end{tabular}
\caption{A summary of classification datasets. }
\end{center}
\end{table}


\begin{table}[h]
\begin{center}
\begin{tabular}{ |c|c|c|c|c| } 
\hline
Dataset & Train Size & Test Size & Features($m$) & Length ($w$)  \\
\hline
ECL&23,377&2,928&321&200  \\ 
Weather&28,005&7,060&12&200  \\ 
ETTm1& 45,697&11,904&7&200 \\ 
\hline
\end{tabular}
\caption{A summary of forecasting datasets. }
\end{center}
\end{table}

\begin{table}[h]
\begin{center}
\begin{tabular}{ |c|c|c|c|c| } 
\hline
Dataset & Train Size & Test Size & Features($m$) & Length ($w$) \\
\hline
SMAP& 135,183&427,617&25&50 \\ 
MSL& 58,317&73,729&55&50 \\ 
SMD& 708,405&708,420&38&200 \\ 
\hline
\end{tabular}
\caption{A summary of anomaly detection datasets. }
\end{center}
\end{table}

\section{Model Size Selection}
We measure model performance as well as computational cost at varying sizes for each model.  To vary the size, we increase the embedding dimension $d$ for each model and dataset combination.  Tables \ref{table:classification_size} and \ref{table:forecasting_size} show the results for each model size and dataset combination.  Overall, we find that the \gls{sbt} generally performs better than the smaller Dense Transformer in terms of performance, except in a few cases.  In all scenarios, the \gls{sbt} model has at least one computational advantage in terms of storage size or FLOPs count.  

Additionally we find that, common with our intuition, datasets with a higher dimensionality $m$ need a higher embedding dimension, while simpler datasets are successful with a smaller embedding dimension.  For example, Insect Wingbeats ($m=200$), Face Detection ($m=144$), and ECL ($m=321$) require $d \ge 128$ to achieve optimal performance.  

\begin{table*}
\large
\begin{center}
\setlength{\tabcolsep}{0.25em}
\begin{tabular}{    ccc|ccc|ccc } 
\toprule
\multicolumn{3}{c}{}&\multicolumn{3}{c}{Dense Transformer}&\multicolumn{3}{c}{Sparse Binary Transformer}\\
\midrule
\multicolumn{1}{c}{Dataset}&\thead{$m$ \\ (Num. Features)} &\multicolumn{1}{c}{$d$}&Accuracy  & \thead{Params (FP32) \\ (K)}&\thead{FLOPs \\ (Mil.)}& Accuracy & \thead{Params (Binary)\\ (K)}&\thead{FLOPs \\ (Mil.)}\\
\hline


 \multirow{4}{*}{Heartbeat}& \multirow{4}{*}{61}
	&	\textbf{64}	&	76.6	&	169.6	&	52.73	&	77.2	&	102.3	&	21.57	\\
&		&	16	&	28.8	&	36.6	&	11.81	&	72.2	&	19.5	&	4.95	\\
&		&	32	&	71.1	&	76.9	&	24.20	&	74.5	&	43.1	&	9.87	\\
&		&	128	&	75.6	&	404.2	&	124.73	&	75.1	&	270.7	&	52.45	\\
\hdashline
 \multirow{4}{*}{Insect Wingbeats}& \multirow{4}{*}{200}
 	&	\textbf{128}	&	63.4	&	555.5	&	5.42	&	64.1	&	420.9	&	2.67	\\
&		&	256	&	65.2	&	1,503.5	&	19.77	&	64.4	&	1,234.9	&	9.80	\\
&		&	64	&	57.9	&	229.0	&	1.60	&	51.4	&	161.3	&	0.79	\\
&		&	400	&	64.9	&	3,040.0	&	46.60	&	64.6	&	2,620.8	&	23.16	\\
\hdashline
 \multirow{4}{*}{Japanese Vowels}& \multirow{4}{*}{12}
	&	\textbf{32}	&	98.0	&	75.5	&	0.33	&	95.3	&	41.6	&	0.16	\\
&		&	8	&	92.5	&	17.7	&	0.05	&	78.5	&	8.9	&	0.02	\\
&		&	16	&	93.6	&	36.0	&	0.12	&	87.7	&	18.8	&	0.06	\\
&		&	64	&	96.2	&	166.9	&	1.01	&	94.1	&	99.9	&	0.49	\\

\hline
\end{tabular}
\caption{Classification Model size selection: Performance of various sized models on each classification dataset.  We include the parameter count as well as FLOPs for both the dense and sparse binary Transformer models.  Parameters are floating-point 32 in the Dense Transformer and Binary in the \gls{sbt}. 
 }\label{table:classification_size}

\end{center}
\end{table*}

\begin{table*}
\large
\begin{center}
\setlength{\tabcolsep}{0.25em}
\renewcommand{\arraystretch}{1.45}%
\begin{tabular}{    ccc|ccc|ccc } 
\toprule
\multicolumn{3}{c}{}&\multicolumn{3}{c}{Dense Transformer}&\multicolumn{3}{c}{Sparse Binary Transformer}\\
\multicolumn{1}{c}{Dataset}&\thead{$m$ \\ (Num. Features)} &\multicolumn{1}{c}{$d$}&MSE  & \thead{Params (FP32) \\ (K)}&\thead{FLOPs \\ (Mil.)}& MSE & \thead{Params (Binary)\\ (K)}&\thead{FLOPs \\ (Mil.)}\\
\hline

 \multirow{4}{*}{ECL}& \multirow{4}{*}{321}
	&	\textbf{350}	&	0.178	&	1,569.4	&	204.76	&	0.187	&	1,569.4	&	74.54	\\
&	&	128	&	0.199	&	348.2	&	40.83	&	0.337	&	345.1	&	10.18	\\
&	&	256	&	0.197	&	956.0	&	120.88	&	0.27	&	951.6	&	40.06	\\
&	&	400	&	0.216	&	1,953.2	&	258.03	&	0.191	&	1,947.2	&	97.21	\\
\hdashline
 \multirow{4}{*}{Weather}& \multirow{4}{*}{7}
 	&	\textbf{100}	&	0.173	&	188.0	&	28.50	&	0.166	&	185.6	&	6.24	\\
& 	&	32	&	0.171	&	44.2	&	6.61	&	0.187	&	42.5	&	0.69	\\
& 	&	64	&	0.172	&	102.7	&	15.53	&	0.177	&	100.6	&	2.61	\\
& 	&	128	&	0.169	&	268.7	&	40.75	&	0.166	&	266.0	&	10.14	\\
\hline
\end{tabular}
\caption{Forecasting Model size selection: Performance of various sized models on each forecasting dataset.  We include the parameter count as well as FLOPs for both the dense and sparse binary Transformer models.  Parameters are floating-point 32 in the Dense Transformer and Binary in the \gls{sbt}. 
 }\label{table:forecasting_size}

\end{center}
\end{table*}



\end{document}